\documentclass[conference]{IEEEtran}
\IEEEoverridecommandlockouts
\usepackage{cite}
\usepackage{amsmath,amssymb,amsfonts}
\usepackage{algorithmic}
\usepackage{graphicx}
\usepackage{textcomp}
\usepackage{xcolor}
\usepackage{tikz}
\usetikzlibrary{positioning}
\usepackage[hidelinks]{hyperref}
\usepackage{float}

\def\BibTeX{{\rm B\kern-.05em{\sc i\kern-.025em b}\kern-.08em
    T\kern-.1667em\lower.7ex\hbox{E}\kern-.125emX}}

\begin{document}

\title{Systematic Literature Review on Clinical Trial Eligibility Matching \\
\thanks{}
}

\author{
    \IEEEauthorblockN{
        \IEEEauthorblockA{
            \begin{minipage}{0.45\textwidth}
                \centering
                \textbf{Muhammad Talha Sharif} \\  
                \textit{Department of Computer Science,}\\ \textit{FAST School of Computing,} \\  
                \textit{National University of Computer}\\ \textit{and Emerging Sciences,} \\  
                Islamabad 44000, Pakistan \\  
                745talha@gmail.com  
            \end{minipage}  
            \hfill  
            \begin{minipage}{0.45\textwidth}
                \centering
                \textbf{Abdul Rehman} \\  
                \textit{Department of Computer Science,}\\ \textit{FAST School of Computing,} \\  
                \textit{National University of Computer}\\ \textit{and Emerging Sciences,} \\  
                Islamabad 44000, Pakistan \\ 
                abdulrehman.ds.pf@gmail.com  
            \end{minipage}
        }
    }
}

\maketitle

\begin{abstract}Clinical trial eligibility matching is a critical yet often labor-intensive and error-prone step in medical research, as it ensures that participants meet precise criteria for safe and reliable study outcomes. Recent advances in Natural Language Processing (NLP) have shown promise in automating and improving this process by rapidly analyzing large volumes of unstructured clinical text and structured electronic health record (EHR) data. In this paper, we present a systematic overview of current NLP methodologies applied to clinical trial eligibility screening, focusing on data sources, annotation practices, machine learning approaches, and real-world implementation challenges. A comprehensive literature search (spanning Google Scholar, Mendeley, and PubMed from 2015 to 2024) yielded high-quality studies, each demonstrating the potential of techniques such as rule-based systems, named entity recognition, contextual embeddings, and ontology-based normalization to enhance patient matching accuracy. While results indicate substantial improvements in screening efficiency and precision, limitations persist regarding data completeness, annotation consistency, and model scalability across diverse clinical domains. The review highlights how explainable AI and standardized ontologies can bolster clinician trust and broaden adoption. Looking ahead, further research into advanced semantic and temporal representations, expanded data integration, and rigorous prospective evaluations is necessary to fully realize the transformative potential of NLP in clinical trial recruitment.
\end{abstract}

\begin{IEEEkeywords}
Clinical Trial Eligibility, Natural Language Processing (NLP), Machine Learning in Healthcare, Medical Text Mining
\end{IEEEkeywords}
\section{Introduction}
Clinical trials are vital not only to the progress of medical science but to ensure new treatment products or interventions are effective as well as safe. Clinical trial eligibility matching begins when identifying potential participants based on inclusion and exclusion criteria devised for the study. Clinical trial eligibility matching ensures that all qualification requirements for medical, demographic, and behavioral qualifications set for participation in the study are met. Accurate eligibility matching increases the validity of trial results, reduces risks for participants, and provides researchers with an opportunity to make valid inferences that could be put to practical use. \\

However, always matching subjects with the correct trials is quite a difficult task. The criteria for clinical trials most of the time refer to some kinds of extended medical history, lab test results, and special characteristics that are unique for each individual trial. Besides, such criteria are mainly presented in natural language, in EHRs or study protocols, thus making it hard to match them through database search. This would result in poor recruitment rates and longer trial durations as well as higher costs. Additionally, the process is staff-intensive and prone to numerous errors by being manual.\\

Artificial Intelligence (AI), and specifically Natural Language Processing (NLP), offers promising solutions to overcome these challenges. NLP can analyze unstructured text in medical records and automatically interpret eligibility criteria, potentially enhancing recruitment speed and accuracy.\\

The capability of NLP methods to parse and interpret eligibility criteria and match them against patient data can help streamline clinical recruitment and make it more accessible to people. More advanced AI techniques can continuously refine and personalize eligibility matching that opens up a way to a more populous variety of people and speeds up the discovery of new treatments. The objective of this study is to provide a systematic overview of the current state of research on clinical trial eligibility matching using NLP by describing methodologies, applications, and how they impact recruitment outcomes.\\

\subsection{\textbf{Objectives}}
This paper attempts to investigate the applications of NLP in increasing the matching of clinical trials with eligibility criteria. Since patient recruitment is viewed as a long staff-intensive process, NLP offers great promise in interpreting complex eligibility criteria, increasing the level of automation in screening, and reducing the healthcare burden on healthcare professionals. The objectives of this paper are to review existing NLP methodologies, review current work up to date with regard to the automation of clinical trial eligibility, and identify challenges and future directions of this research field. Here are research objectives and their corresponding questions.
\\

\textbf{Objective 1: Review NLP methodologies and techniques for Clinical Trial Eligibility.}
\begin{itemize}
    \item Q1: What NLP techniques are used for Clinical Trial Eligibility?
    \item Q2: How effective are these NLP techniques in evaluating Clinical Trial Eligibility?
\end{itemize}

\textbf{Objective 2: To review the best methodologies for automating Clinical Trial eligibility determination.}
\begin{itemize}
    \item Q1: What is the best state-of-the-art NLP technology used for automating Clinical Trial eligibility?
\end{itemize}

\textbf{Objective 3: To explore challenges and future directions in automating the matching of Clinical Trial eligibility criteria.
}

\begin{itemize}
    \item Q1: What challenges are associated with automating Clinical Trial eligibility criteria matching?
    \item Q2: What are the potential future directions for enhancing eligibility criteria automation?
\end{itemize}

\subsection{\textbf{Literature Search Strategy}}
To find relevant literature on "Clinical Trial Eligibility Using NLP," a systematic search of Google Scholar and Mendeley was conducted, using filters and relevance criteria to narrow the search down to only the most relevant studies.\\

\begin{figure}[h]
    \centering
    \includegraphics[width=0.5\textwidth]{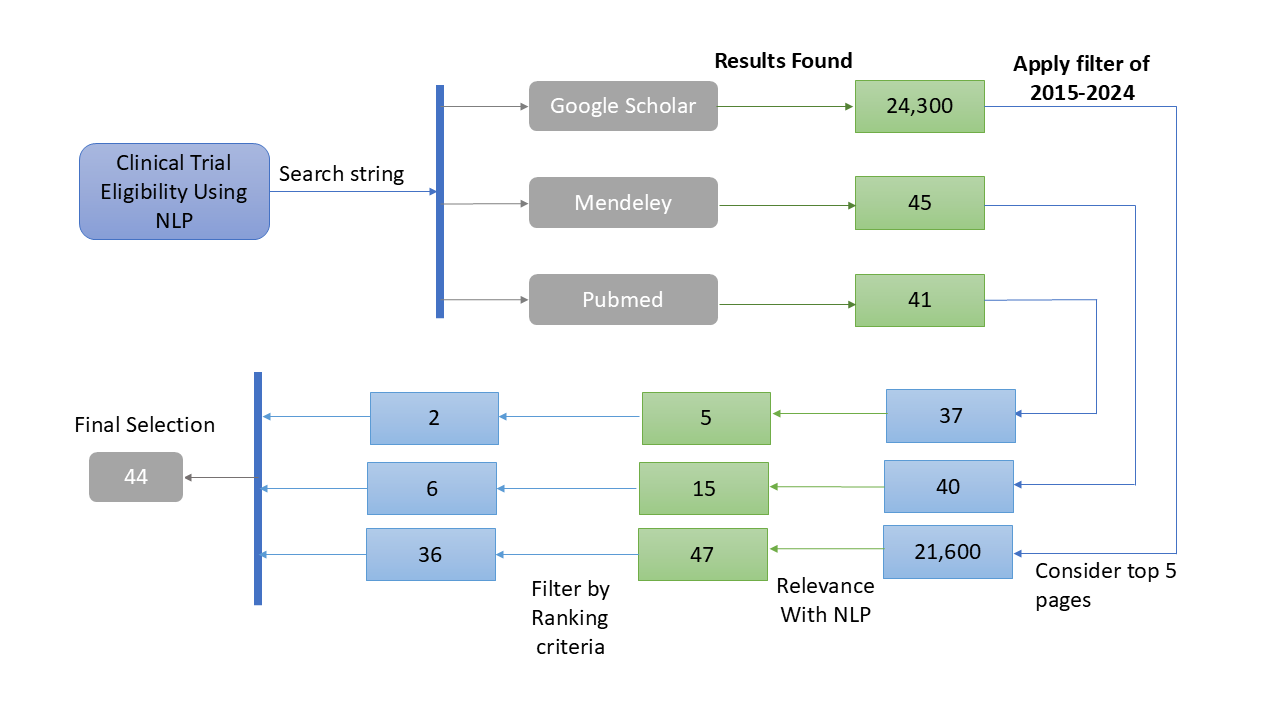} 
    \caption{Research paper selection criteria}
    \label{fig:example}
\end{figure}

\subsubsection{\textbf{Google Scholar Search}}
Doing an initial Google Scholar search with the query “Clinical Trial Eligibility Using NLP" elicited around 24,300 results. Narrowing down the selection to the research conducted in recent times from 2015 to 2024 cut down results to 21,600. From there, the following analysis was conducted on the first ten pages of ten records each, containing the studies most relevant to the task at person matches for clinical trial matching and NLP methodologies. This screen elicited 47 relevant studies.
More stringent filtering criteria to ensure academic rigour applied. For the case of journal-based studies, their selection was based on a Journal Prestige Index (JPI) of greater than 60 and W category ranking, whilst conference papers had to have an A or A* rating. After these filters, 36 quality papers remains.\\

\subsubsection{\textbf{Mendeley Search}}
A further search on Mendeley using the same keywords returned a further 45 results. Adding the date range filter from 2015 to 2024 reduced the number further to 40. Then, each paper was filtered by relevance to how NLP techniques can be applied in eligibility for clinical trials. This produced 15 relevant papers.
Using almost the same inclusion criteria as with the Google Scholar search, namely journals with a JPI above 60 and having a W category rating or A/A* conference papers, six papers were found.\\

\subsubsection{\textbf{Pubmed Search}}
A further search on Pubmed using the same keywords returned a further 41 results. Adding the date range filter from 2015 to 2024 reduced the number further to 37. Then, each paper was filtered by relevance to how NLP techniques can be applied in eligibility for clinical trials. This produced 5 non duplicate and relevant papers as 7 papers found there was duplicate by previous papers. Using almost the same inclusion criteria as with the Google Scholar search, namely journals with a JPI above 60 and having a W category rating or A/A* conference papers, 2 papers were found.\\

\subsubsection{\textbf{Final Selection}}
Following the combination of the filtered results across both search platforms, 44 high-quality, relevant papers were chosen to be used for the review. Being a curated set, it represents the back-bone of this literature review and hence presents a comprehensive overview of present methodologies and challenges and future directions in clinical trial eligibility matching through NLP.
\begin{table*}[htbp]
\caption{Selected papers details}
    \begin{tabular}{|p{170pt} |p{40pt}| p{40pt} |p{95pt} |p{60pt}| p{40pt} |p{35pt}|}

\hline
    \centering
        \textbf{Name} & \textbf{Type} & \textbf{Rank} & \textbf{venue} & \textbf{Search from} & \textbf{publish date} \\ \hline
        Automated clinical trial eligibility prescreening: increasing the efficiency of patient identification for clinical trials in the emergency department & Journal & 84 W & Journal of the American Medical Informatics Association : JAMIA & google scholor & 16-Jul-14 \\ \hline
        Increasing the efficiency of trial-patient matching: automated clinical trial eligibility Pre-screening for pediatric oncology patients & Journal & 78 W & BMC Medical Informatics and Decision Making & google scholor & 14-Apr-15 \\ \hline
        Artificial Intelligence Tool for Optimizing Eligibility Screening for Clinical Trials in a Large Community Cancer Center & Journal & 65 W & JCO clinical cancer informatics & google scholor & 24-Jan-20 \\ \hline
        Automated classification of eligibility criteria in clinical trials to facilitate patient-trial matching for specific patient populations & Journal & 84 W & Journal of the American Medical Informatics Association : JAMIA & google scholor & 19-Feb-17 \\ \hline
        EliIE: An open-source information extraction system for clinical trial eligibility criteria & Journal & 84 W & Journal of the American Medical Informatics Association : JAMIA & google scholor & 1-Apr-17 \\ \hline
        Natural Language Processing for Mimicking Clinical Trial Recruitment in Critical Care: A Semi-Automated Simulation Based on the LeoPARDS Trial & Journal & 81 W & IEEE Journal of Biomedical and Health Informatics & google scholor & 9-Mar-20 \\ \hline
        Clinical trial cohort selection based on multi-level rule-based natural language processing system & Journal & 84 W & Journal of the American Medical Informatics Association : JAMIA & google scholor & 13-Jul-19 \\ \hline
        DeepEnroll: Patient-Trial Matching with Deep Embedding and Entailment Prediction & Conference & A* & International World Wide Web Conference & google scholor & 20-Apr-20 \\ \hline
        Case-based reasoning using electronic health records efficiently identifies eligible patients for clinical trials & Journal & 84 W & Journal of the American Medical Informatics Association : JAMIA & google scholor & 13-Mar-15 \\ \hline
        Identifying Patient Phenotype Cohorts Using Prehospital Electronic Health Record Data & Journal & 62 W & Prehospital Emergency Care & Mendley & 1-Dec-20 \\ \hline

 Representing cancer clinical trial criteria and attributes using ontologies: An NLP-assisted approach. & Journal & 99 W & Journal of Clinical Oncology & Mendley & 25-May-20 \\ \hline
        Phenotyping of clinical trial eligibility text from cancer studies into computable criteria in electronic health records. & Journal & 99 W & Journal of Clinical Oncology & Mendley & 28-May-21 \\ \hline
        Cognitive technology addressing optimal cancer clinical trial matching and protocol feasibility in a community cancer practice. & Journal & 99 W & Journal of Clinical Oncology & Mendley & 30-May-17 \\ \hline
        A natural language processing tool for automatic identification of new disease and disease progression: Parsing text in multi-institutional radiology reports to facilitate clinical trial eligibility screening. & Journal & 99 W & Journal of Clinical Oncology & Mendley & 28-May-21 \\ \hline
        The Leaf Clinical Trials Corpus: a new resource for query generation from clinical trial eligibility criteria & Journal & 95 W & Scientific data & Mendley & 11-Aug-22 \\ \hline
        Piloting an automated clinical trial eligibility surveillance and provider alert system based on artificial intelligence and standard data models & Journal & 79 W & BMC Medical Research Methodology & Pubmed & 11-Apr-23 \\ \hline
        EMR2vec: Bridging the gap between patient data and clinical trial & Journal & 93 W & Computers and Industrial Engineering & Pubmed & 15-Mar-21 \\ \hline
        Automated Matching of Patients to Clinical Trials: A Patient-Centric Natural Language Processing Approach for Pediatric Leukemia & Journal & 65 W & JCO clinical cancer informatics & Google scholor & 10-Jul-23 \\ \hline
        Matching patients to clinical trials with large language models & Journal & 98 W & Nature Communications & Google scholor & 18-Nov-24 \\ \hline
 A review of research on eligibility criteria for clinical trials  & Journal & 69 W & Clinical and Experimental Medicine & Google scholar  & 05-Jan-2023 \\ \hline
        Quantifying risk associated with clinical trial termination: A text mining approach & Journal & 86 W & Information Processing and Management & Google scholar  & May 2019 \\ \hline
        Cohort Selection for Clinical Trials From Longitudinal Patient Records: Text Mining Approach& Journal & 73 W & JMIR Medical Informatics & Google scholar  & 31-Oct-2019 \\ \hline
        Automated classification of clinical trial eligibility criteria text based on ensemble learning and metric learning & Journal & 78 W &  BMC Medical Informatics and Decision Making  & Google scholar  & 30-07-2021 \\ \hline
        Analysis of eligibility criteria clusters based on large language models for clinical trial design & Journal  & 84 W & Journal of the American Medical Informatics Association & Google scholar  & 26 Dec 2024 \\ \hline
        Predicting Publication of Clinical Trials Using Structured and Unstructured Data: Model Development and Validation Study & Journal & 73 W & JMIR Medical Informatics & Google scholar  & 19 Apr 2022 \\ \hline
       
    \end{tabular}
\end{table*}

\begin{table*}[htbp]
    \begin{tabular}{|p{170pt} |p{40pt}| p{40pt} |p{95pt} |p{60pt}| p{40pt} |p{35pt}|}
    \hline
     A comparative study of pre-trained language models for named entity recognition in clinical trial eligibility criteria from multiple corpora & Journal  & 78 W &  BMC Medical Informatics and Decision Making  & Google scholar  & 06 Sep 2022 \\ \hline
   Key indicators of phase transition for clinical trials through machine learning& Journal  & 93 W & Drug Discovery Today & Google scholar  & Feb 2020 \\ \hline
        PRISM: Patient Records Interpretation for Semantic clinical trial Matching system using large language models  & Journal  & 95 W & NPJ Digital Medicine & Google scholar  & 28 Oct 2024 \\ \hline
        Prediction of Clinical Trials Outcomes Based on Target Choice and Clinical Trial Design with Multi-Modal Artificial Intelligence  & Journal  & 94 W & Clinical Pharmacology \& Therapeutics & Google scholar  & 22 July 2023 \\ \hline
        Evaluating eligibility criteria of oncology trials using real-world data and AI  & Journal  & 100 W & Nature Publishing Group & Google scholar  & 07 April 2021 \\ \hline
  Fine-Tuned Transformers and Large Language Models for Entity Recognition in Complex Eligibility Criteria for Clinical Trials & Conference & A & International Conference on Information Systems Development 
 & Google scholar  & 09 Sep 2024 \\ \hline
        A clinical trials corpus annotated with UMLS entities to enhance the access to evidence-based medicine & Journal  & 78 W &  BMC Medical Informatics and Decision Making  & Google scholar  & 22 Feb 2021 \\ \hline
        An Ensemble Learning Strategy for Eligibility Criteria Text Classification for Clinical Trial Recruitment: Algorithm Development and Validation & Journal  & 73 W & JMIR Medical Informatics & Google scholar  & July 2020 \\ \hline
        ELaPro, a LOINC-mapped core dataset for top laboratory procedures of eligibility screening for clinical trials & Journal  & 78 W &  BMC Medical Informatics and Decision Making  & Google scholar  & 14 May 2022 \\ \hline
        Artificial Intelligence in Cardiovascular Clinical Trials& Journal  & 89 W & Journal of the American College of Cardiology & Google scholar  & Nov 2024 \\ \hline
        Application of Information Technology to Clinical Trial Evaluation and Enrollment  & Journal  & 96 W & JAMA Oncology & Google scholar  & 08 July 2021 \\ \hline
        Accuracy of an Artificial Intelligence System for Cancer Clinical Trial Eligibility Screening: Retrospective Pilot Study & Journal  & 73 W & JMIR Medical Informatics & Google scholar  & Mar 2021 \\ \hline
        Combining human and machine intelligence for clinical trial eligibility querying& Journal  & 84 W & Journal of the American Medical Informatics Association & Google scholar  & July 2022 \\ \hline
        Machine learning and natural language processing in clinical trial eligibility criteria parsing: a scoping review & Journal  & 93 W & Drug Discovery Today & Google scholar  & Oct 2024 \\ \hline
        A Real-Time Automated Patient Screening System for Clinical Trials Eligibility in an Emergency Department: Design and Evaluation & Journal  & 73 W & JMIR Medical Informatics & Google scholar  & 28 Mar 2019 \\ \hline
        AutoCriteria: a generalizable clinical trial eligibility criteria extraction system powered by large language models & Journal  & 84 W & Journal of the American Medical Informatics Association & Google scholar  & 11 Nov 2023 \\ \hline
        Learning Eligibility in Cancer Clinical Trials Using Deep Neural Networks  & Journal  & 73 W & Applied Sciences & Google scholor & 23 July 2018 \\ \hline
\end{tabular}
\end{table*}

\section{Dataset Details}

\cite{b3} paper refers to dataset that includes a total of 202795 patient encounters (the gold standard subset covers 600 encounters over 13 trials), obtained from Cincinnati Children’s Hospital Medical Center. This information consists of both structured data (e.g. demographics, laboratory results) and unstructured data such as clinical notes, albeit with varying levels of unstructured data availability. while \cite{ni2015increasing} paper data set comprises 215 pediatric oncology patients, 55 studies, and 169 patient-trial combinations, with the inclusion criteria obtained from ClinicalTrials.gov and the patient information from EHRs of the Cincinnati Children’s Hospital Medical Center. The focus is on the area of pediatric cancer and related diseases and ensures data integrity through validated time markers and uniform protocols of enrollment.\\

\cite{beck2020artificial} comprised dataset of roughly 11,000 patients with breast malignancy, incorporating inclusion criteria from ClinicalTrials.gov, trial protocols by Novartis, and EHRs from Highlands Oncology Group. It utilizes formalized procedures of NLP and tailored breast cancer patient data models to ensure high-quality eligibility criteria screening. \cite{zhang2017automated} paper dataset consists of 891 clinical studies which are related to cancer and HIV and are available at ClinicalTrials.gov. It has primary data such as study considerations and segmented eligibility threshold requirements which were tagged out using the regular expressions methods for precision.\\

\cite{b9} data has a time span data of up to 4.5 million patients out of which 30,000 were sampled for the analysis and 262 participants were sampled from clinical trials on Type 2 diabetes, HIV and multiple myeloma. The data was also of high-quality as it was collated with proper preprocessing and systematic relevance scoring by the aforementioned researchers.\\

paper \cite{b15} Leaf Clinical Trials Corpus data set comprises 804 training and 202 testing samples with demographical information such as age, ethnicity, life stage, and family ties. It contains high quality annotations that have gone through several rounds of editing aimed at ensuring that there is semantic coherence through interconnected entities and values.

\begin{table*}[htbp]
\begin{tabular}{|p{100pt} |p{40pt}| p{50pt} |p{90pt} |p{60pt}| p{100pt}|}

      \hline
        \textbf{Paper} & \textbf{Size} & \textbf{Demographics} & \textbf{Source} & \textbf{Disease Diversity} & \textbf{Data Quality} \\ \hline
        Automated clinical trial eligibility prescreening: increasing the efficiency of patient identification for clinical trials in the emergency department & - Gold standard: 600 encounters, 13 trials. 
- Reference dataset: 202,795 encounters. & Patient's age, gender, Language & Cincinnati Children’s Hospital Medical Center & ~ & Includes structured fields (demographics, lab data) and unstructured fields (diagnosis, clinical notes). Not all encounters had complete unstructured data. \\ \hline
        Increasing the efficiency of trial-patient matching: automated clinical trial eligibility Pre-screening for pediatric oncology patients & - 55 trials. 
- 215 patients. 
- 169 patient-trial matches. & Age, Gender & ClinicalTrials.gov for eligibility criteria of trials; patient data from CCHMC (Cincinnati Children's Hospital Medical Center) EHR records. & Cancer-related diseases; includes specific diagnoses and ICD-9 codes representing a variety of clinical problems. & Diagnoses and notes timestamps validated to exclude post-enrollment or post-trial-closing data. 
- More descriptive eligibility criteria compared to earlier studies. 
- High consistency in pediatric oncology enrollment.\\  \hline
        Artificial Intelligence Tool for Optimizing Eligibility Screening for Clinical Trials in a Large Community Cancer Center & approximately 11,000 & Age, Gender & - Trial Data: Extracted from ClinicalTrials.gov (e.g., NCT01633060, NCT02437318, NCT02422615, NCT01923168) and trial protocols provided by Novartis Pharmaceuticals. 
- Patient Data: Electronic health records (EHRs) from Highlands Oncology Group. & breast cancer & - Eligibility criteria manually reviewed to validate WCTM’s NLP interpretations. 
- Data processed through traditional NLP steps (e.g., tokenization, negation detection). 
- Patient data models customized for breast cancer\\ \hline
        Automated classification of eligibility criteria in clinical trials to facilitate patient-trial matching for specific patient populations & 891 & Not mentioned & ClinicalTrials.gov & Cancer and HIV & - Structured Data: Study titles, conditions studied, eligibility criteria segmented into inclusion and exclusion. 
- Regular expressions refined with specificity prioritization used for annotation \\ \hline
EliIE An open-source information extraction system for clinical trial eligibility criteria & 8008 & Not mentioned &  ClinicalTrials.gov & Alzheimer & Entity and relation annotation validated by a clinician and informatics student. Relations of atributes are well defined 
 \\ \hline
        Case-based reasoning using electronic health records efficiently identifies eligible patients for clinical trials & - Clinical Trial Participants: 4–128 participants per trial (total 262). 
- General Patient Data: 30,000 randomly selected patients. 
- Data warehouse contains longitudinal records for \~4.5 million patients. & Not mentioned & - Columbia University Clinical Data Warehouse. 
- Electronic Health Records. & Focus on Type 2 diabetes, HIV, and multiple myeloma & Overall Data Quality and annotation quality was ensured by researchers  through preprocessing and systematic relevance scoring. \\ \hline
        The Leaf Clinical Trials Corpus: a new resource for query generation from clinical trial eligibility criteria & 804 training 202 testing & age, ethnicity, life stage, and family relationships. & Leaf Clinical Trials (LCT) corpus & preeclampsia and some other diseases(not explicitly mentioned) & High-quality annotations developed through iterative refinement based on task requirements. 
- Semantic consistency ensured via explicit entity-value relationships and relations. \\ \hline
        Piloting an automated clinical trial eligibility surveillance and provider alert system based on artificial intelligence and standard data models & 21,974 clinical text notes from 400 patients & Not mentioned & Clinical text notes derived from EHR at the Medical University of South Carolina. & Includes patients from cardiovascular and cancer trials. & High-level quality ensured with detailed annotation for a small subset. 
The remaining data (21,954 notes) lacks detailed annotation information. \\ \hline
        EMR2vec: Bridging the gap between patient data and clinical trial & 5000 after cleaning remains 1500 & Not mentioned & SNOMED-CT ontology for medical term normalization. 
Electronic Medical Records & Stroke, Osteoarthritis, Thyroid Cancer, Prostate Cancer, Breast Cancer and Obesity & Data is normalized using SNOMED-CT for consistency. 
Entity extraction uses Named Entity Recognition to ensure structured representation. \\ \hline
       
    \end{tabular}
\end{table*}

\begin{table*}[htbp]
\begin{tabular}{|p{100pt} |p{40pt}| p{50pt} |p{90pt} |p{60pt}| p{100pt}|}

      \hline
 Automated Matching of Patients to Clinical Trials: A Patient-Centric Natural Language Processing Approach for Pediatric Leukemia & 20 Patients & children and adolescents & ClinicalTrials.gov 
Synthetic Cohort & leukemia & Structured XML files ensure high-quality input. 
Text preprocessing included tokenization, lemmatization, and removal of noise. 
Validation with synthetic cohort. \\ \hline
        Matching patients to clinical trials with large language models & 183 Patients records from Synthetic Cohort 
23000 records from TREC Cohorts  & Not mentioned & Synthetic Dataset 
TREC CT Tracks  & Multiple disease including Cancer & Three eligibility labels used irrelevant, potential, eligible for data quality. 
Expert evaluations and LLM-driven insights improve annotation consistency. \\ \hline

    \end{tabular}
\end{table*}

\section{EHR Data Integration}

As a matter of fact, the research designed LCFs for encounters deemed ineligible by using standard EHR fields (Eg. demographic, lab data). Formal clinical terminologies and other unstructured information were extracted from notes and diagnoses via NLP Techniques. Creating patterns of encounters involved assembling vectors from clinical notes for unstructured fields and developing trial pattern vectors from trial criteria comprises of terms from previously successful patients in order to get hypernyms. Patients were represented using both LCFs and vectorized unstructured data, which contained demographic information, lab data, diagnosis, and clinical notes.\cite{b3}

Preprocessing comprised filtering patients by demographics such as age and gender, removing all diagnoses and notes which date after the enrollment, and applying enrollment trial windows to exclude patients with clinical notes. The feature extraction stage used advanced NLP tools including: Apache cTAKES, UMLS for mapping ICD-9 codes to SNOMED-CT terms, NegEx algorithm for negation handling, and extraction of Concept Unique Identifiers from the Medical Language System. Structured data, which included ICD-9 codes and demographics, as well as unstructured data, including clinical notes, were used to represent patients in which terms that were extracted were stored as bag-of-words in patient vectors containing notes, diagnoses and mapped codes.\cite{ni2015increasing}

paper \cite{beck2020artificial} the privatization of NLP would be needed by extracting eligibility and exclusion criterions from ClinicalTrials.gov and trial protocols, processing unstructured EHR data such as clinical and progress notes, and trial filters to lower the patient-reviews workload. Feature extraction designated medical attributes such as laboratory tests, cancer diagnoses, sex, and age using NLP techniques like tokenization, POS tagging, and detection of negations and hypotheticals. Patients were represented through structured data such as: demographics, lab tests and unstructured data such as: progress notes, with relevant attributes mapped to trial criteria for efficient matching.

\begin{table*}[htbp]
\label{tabehr}
\caption{EHR Data table}
    \begin{tabular}{|p{120pt} |p{100pt}| p{100pt} |p{100pt}|}
    \hline
   \textbf{ Paper Name} & \textbf{Preprocessing methods} & \textbf{Feature Extraction} & \textbf{Patient Representattion} \\ \hline
        Automated clinical trial eligibility prescreening: increasing the efficiency of patient identification for clinical trials in the emergency department & - Structured EHR fields (e.g., demographics, lab data) were used to create Logical Constraint Filters (LCFs) for excluding ineligible encounters.
- Unstructured fields (e.g., clinical notes, diagnosis) were processed using NLP to extract medical terms. & - Medical terms were extracted from unstructured fields and stored in encounter pattern vectors.
- Trial pattern vectors were constructed using trial criteria and extended with patterns from EHRs of previously eligible patients (to capture hyponyms). & - Patients were represented using  EHR data fields categorized as: 
1. Structured fields (e.g., demographics, lab data). 
2. Unstructured fields (e.g., diagnosis, clinical notes). 
- Representation included both structured LCFs and vectorized unstructured data. \\ \hline
        Increasing the efficiency of trial-patient matching: automated clinical trial eligibility Pre-screening for pediatric oncology patients & Applied a demographics-based filter (age, gender). 
Excluded diagnoses and notes entered after a trial’s enrollment. 
Applied trial enrollment windows to filter patients who lacked clinical notes. & Used advanced NLP such as  Apache and  cTAKES 
Mapped ICD-9 codes to SNOMED-CT terms using UMLS. 
Processed negations like ""No CSN disease"" using the NegEx algorithm. 
Extracted Concept Unique Identifiers from Medical Language System. & Represented patients using structured data include demographics, ICD-9 codes 
Represented patients unstructured data like clinical notes. 
Stored extracted terms in patient vectors as a bag-of-words. 
Patient vectors included terms from clinical notes, diagnoses, and mapped codes.\\ \hline
        Artificial Intelligence Tool for Optimizing Eligibility Screening for Clinical Trials in a Large Community Cancer Center & NLP was used to process unstructured data from EHRs such as clinical notes and progress notes.
Eligibility and exclusion criteria were extracted from ClinicalTrials.gov and trial protocols using an NLP pipeline. 
Trial filters were added to reduce the patient review workload. & Extracted medical attributes from patient EHR data using NLP annotators to identify named entities and relationships between entities. 
Included attributes such as laboratory tests, cancer diagnosis, sex, and age.
Used tokenization, parts-of-speech tagging, negation, and hypotheticals detection and normalization. & Represented patients using structured data like demographics, laboratory test. 
Represented patients unstructured data progress notes. 
Populated patient data models with attributes relevant to criteria. 
Patient attributes were mapped to trial criteria for matching.\\ \hline
        EliIE: An open-source information extraction system for clinical trial eligibility criteria & Normalized punctuation in the eligibility criteria text. 
Removed criteria not available in EHR, such as informed consent or patient willingness to participate. 
Keywords like ""informed consent"" were used to filter nonapplicable criteria. & Entity and Attribute Recognition: Used CRF-based sequence labeling to extract entities (condition, observation, drug/substance, procedure/device) and attributes (measurement, temporal constraint, qualifier/modifier). 
Features included word-level features, part-of-speech tags, and lemmas using NLTK. 
UMLS-based features identified concepts in EC text. 
Implemented semantic word representations using techniques like Brown clustering and Word2Vec. & The study used concept normalization to standardize entities and attributes using the OMOP Common Data Model (CDM) vocabulary
The output was structured in XML format for downstream large-scale analysis.
\\ \hline
        Case-based reasoning using electronic health records efficiently identifies eligible patients for clinical trials & Data types extracted from EHR these are medication orders, diagnosis, laboratory results, and free-text clinical notes. 
Data normalization via the Medical Entities Dictionary (MED), mapping concepts to latest RxNorm. 
Free-text clinical notes processed with named entity recognition (NER), semantically similar concept aggregation, negation detection, redundancy handling, and topic modeling. & Medications and Diagnosis done by researchers in which resarchers Counts of normalized MED-based codes. 
Laboratory Results are geather by taking average values for numerical tests, or presence noted for categorical results (e.g., “positive” or “negative,” “high” or “low,” or a free-text summary). 
Clinical Notes extracted by precessing using named entity recognition, semantically similar concepts aggregation, negation detection, redundancy handling, and topic modeling. & Each patient represented by four vectors: medication codes, diagnosis codes, lab results, and clinical note topics. 
The ""target patient"" profile for each trial is aggregated by analyzing common concepts from trial participants' data and summarizing them into vectors. 
Similarity between new patient and ""target patient"" calculated using cosine similarity, with aggregation across data types to generate a final relevance score. \\ \hline
\end{tabular}
\end{table*}

The preprocessing involved normalizing punctuation in eligibility criteria, removing non-EHR-applicable criteria (e.g., informed consent), and filtering out irrelevant criteria using keywords. Feature extraction utilized CRF-based sequence labeling to identify entities (e.g., conditions, drugs, procedures) and attributes (e.g., measurements, temporal constraints) with features like word-level data, POS tags, lemmas, and UMLS-based concept mapping. Semantic word representations, including Brown clustering and Word2Vec, were implemented for deeper understanding. Patient representation employed concept normalization with the OMOP Common Data Model vocabulary, outputting structured XML for large-scale analysis.\cite{kang2017eliie}

Pre-processing of EHR data types included medication orders, diagnoses, lab tests, and free-text clinical notes, which were normalized using the Medical Entities Dictionary by matching concepts to RxNorm for further processing. Free-text clinical notes involved processing such as Named Entity Recognition, semantic concept aggregation, negation detection, redundancy handling, and topic modeling. The feature extraction process included counting normalized MED-based codes for medications and diagnoses, averaging numerical lab results, noting categorical outcomes, and extraction of clinical note topics. Four vectors representing the patient were used: medication codes, diagnosis codes, lab results, and clinical note topics, whereas the "target patient" profiles were aggregated from data of trial participants in order for the similarity to be calculated using cosine similarity to yield a final relevance score.

In paper \cite{b9} By pre-processing EHR data types: Medication orders, diagnoses, lab results, and clinical notes were free textnormalized to the Medical Entities Dictionary (MED), which maps concepts to RxNorm for any further processing. Free text clinical notes processes named entity recognition, semantic concept aggregation, negation detection, redundancy handling, topic modeling, etc. Feature extraction this counts normalized results of MED-based codes for medications and diagnosis, averages numerical lab results and records categorical outcomes, with extraction of clinical note topics. Four vectors represented the patient: medication codes, diagnosis codes, lab results, and clinical note topics, while "target patient" profiles were aggregated from data of trial participants for similarity calculation using cosine similarity to yield a final relevance score.

\section{Discussion}
Clinical trial eligibility matching has been significantly evolved with the integration of Natural Language Processing and machine learning techniques. Various techniques are applied to automate and expedite this process.
\subsection{Advancements in Data Utilization}
Among the cornerstone achievements in this respect, under clinical trial eligibility systems, it is effective to make use of both structured and unstructured data. Most of the studies utilize the electronic health record, which has both structured fields like patient demographics, lab results and unstructured fields, including clinical notes and diagnostic summaries. The combination of these different types of data has helped the system achieve a higher level of accuracy.
Similarly, disease-specific datasets show the advantage of domain-specific data in enhancing recall and precision. However, most of the studies were troubled by the problem of data insufficiency or imbalance, particularly among smaller cohorts. The development and integration of synthetic datasets, including TREC Tracks and synthetic cohorts, have provided a promising solution to overcome such limitations and therefore allow for powerful model testing even in underrepresented scenarios.

\subsection{Quality of Annotations and Validation}
High-quality annotations are very importance to the success of clinical trial matching systems. EliIE and Leaf Clinical Trials Corpus are examples of studies that have shown great care in annotation.
\textbf{EliIE} \cite{kang2017eliie} validated its entity and relation annotations via expert reviews; it achieved an F1-score of 0.89 for relation extraction and 0.79 for entity recognition. All the more, because of this stringent approach, extracted attributes could precisely represent eligibility criteria.
\textbf{The Leaf Clinical Trials Corpus} \cite{b15} used an iterative refinement in the curation process to ensure semantic consistency, with precision, recall, and F1 scores all above 80\%.
These two approaches can be considered the most efficient in all given papers and can be used in future as well. Manual and semi-automated validation methods increase the reliability of annotations. However, these approaches are very resource-intensive, which brings up the need for automated systems that could reach the quality of manual annotations without losing efficiency.

\subsection{Methodological Innovations}
NLP and machine learning methodologies have greatly pushed the frontier of automating patient-trial matching. Some important innovations include:\\
\textbf{Rule-Based and Machine Learning Models:} The traditional rule-based systems, for example, used in Automated Classification of Eligibility Criteria \cite{zhang2017automated}, are now complemented with machine learning approaches such as NER and bag-of-words models. These hybrid systems achieved a macro-averaged F2 score of 0.87, thus demonstrating the synergy between structured rule-based logic and machine learning techniques.\\
\textbf{Contextual Embeddings and Ontology-Based Normalization:} Most recent studies start to take advantage of contextual embeddings such as BERT in DeepEnroll \cite{b8} and ontology-based normalization such as SNOMED-CT in EMR2vec \cite{b17} in an attempt to handle the semantic complexity of eligibility criteria. Such methods guarantee consistency across datasets and improve interpretability.\\
\textbf{Explainability in AI Models:} The explainable models, such as Cognitive Technology for Cancer Trial Matching \cite{b13}, facilitate clear reasons for excluding patients. Clinicians find this more trustworthy, and higher adoption is realized.
The majority of the existing systems are still text-based data-based, but there is an unexplored opportunity in integrating genomic, imaging, and wearable data for developing more comprehensive eligibility profiles.\\
These represent the main techniques that will play a prime role in future automated trail matching.

\section{Limitations in selected papers}
The selected papers have some limitations that are discuss further, \cite{b3} has some limitations and areas for improvement that can be identified in the paper on automated ES for clinical trials. Data availability is considered a critical challenge because the approach depends very much on the completeness of EHR data fields, some of which were missing, like gestational age, causing errors in patient recommendations. Further the bag-of-words approach used in the NLP module limits the ability to capture semantic and temporal relationships within clinical narratives, leading to false positives, especially when signs and symptoms are alike or when exclusions are just implied. In this paper \cite{ni2015increasing} researchers analyzed the error of the ES algorithm, which found out major flaws and shortcomings of this algorithm. It further reported that 54.7\% of errors occurred from the algorithm's failure to discriminate between the similar medical terms used: "T cell lymphoblastic lymphoma" and "Pre-B cell lymphoblastic lymphoma." Since this algorithm was based on words instead of semantic relations, this will be the area where the further improvement may be done with more advanced NLP techniques on the concept recognition. Moreover, the algorithm had trouble distinguishing between new and historical diagnoses, which necessitates temporal reasoning. The logic-based filter, limited to demographics, missed important exclusion criteria, such as prior enrollment.\\

Some of the limitations of study \cite{beck2020artificial} include its small sample size, its only institution was a rural one, and only breast cancer trials, thus limiting generalizability to other settings, cancers, or types of trials. Moreover, the model ability to extract attributes and thus to make determinations about trial eligibility from only one medical note may be incomplete. Using all clinical data as in WCTM-EHR integration, it would most likely increase system performance. The manual review was performed by just one coordinator, and the timing study occurred on only one day, which reduces the scope of the study. The study \cite{zhang2017automated} has limitations, like the segmentation of inclusion and exclusion criteria, which are reliably challenging. Besides, the complexity of adding context might have helped, but it was left out. Future work may look into more advance methods in text processing, using sentence segmentation and part-of-speech tagging for higher precision. Furthermore, the reliance on SVM-based classifiers means a lot of upfront annotation, which is fine for the research team but might limit scalability for larger, evolving problems. This study was done retrospectively and hence limited in how it applies to real-world scenarios. A few of the limitations for the study \cite{kang2017eliie} of the EliIE system are that annotator agreement was not reached because varying levels of granularity of concept can mess up the consistency and quality of the annotated data. Moreover, its generalizability is limited since it's not tested outside Alzheimer's Disease (AD) and neuropsychological diseases. Diseases in other regions may differ in some aspects, such as anatomical locations and genetic information, from AD. Moreover, the query model adopted here only handles one clinical entity at a time and cannot handle more complex queries involving more than one entity, such as disease-drug combinations.\\

The methodology doesn't apply well in this paper \cite{b9}, more complex trial designs with multiple cohorts by simplifying the problem to clinical trials that have a single patient cohort. Such a study used basic techniques to avoid bias, but more advanced methods could improve the results. For instance, it would bring temporal trends from lab results into patient EHRs and model diagnoses and medications with probability distributions, hence filling data gaps and improving identification of eligible patients. The "target patient" model might also profit from statistical methods like mixture models or hidden Markov models for better predictions. Also, the study depended on simple concept-based reasoning with UMLS and MED, so integrating more medical ontologies might boost search precision and recall. The LCT corpus used in paper \cite{b15} has a few limitations, which may impact its performance and generalizability: it is mostly singly annotated, with only 11\% of the documents double annotated, which could hurt consistency and reliability of the annotations; nearly half of the entities have been auto-predicted and then manually corrected which may introduce bias if checked less thoroughly. Although experiments reveal high F1 scores and small differences between performance for manually and semi-automatically annotated data, automatic prediction still could be a problem. The second limitation is the fact that formal queries are missing from the corpus; thus, the task is challenging to determine how well the annotations would work for query generation or to compare it to other corpora. 

\section{Conclusion}
Integration of NLP and machine learning in clinical trial eligibility matching has dramatically evolved the workflow, changing what has been traditionally manual and time-intensive into an increasingly automated, scalable, and precise workflow. While there are still some significant challenges to overcome, especially concerning data completeness and domain scalability. But advanced techniques and new machine learning algorithms include further advances in NLP techniques, methods of annotation, and synthetic datasets to allow clinical trials to more easily reach the appropriate patients and, thereby, accelerate medical research and advance patient care.
Addressing the highlighted challenges and leveraging methodological advancements, researchers and practitioners can build systems that not only enhance clinical trial efficiency but also democratize access to cutting edge treatments for diverse patient populations worldwide.

\section{Future Directions of selected papers}
The future Works of the selected papers and their future directions are discussed here. To better interpret models and improve their accuracy, further work might require in given study \cite{b3} to incorporation of more expansive EHR data fields and utilization of advanced NLP algorithms that are based on semantic and temporal context, for example, bag-of-phrases approaches. The other thing would be to extend testing and validation to a more generalized environment of prospective, randomized controls and patients in diverse settings across several institutions, which would possibly help to allay the generalizability issues. These may enhance the resilience as well as the scalability of the ES system toward a wider application. For paper \cite{ni2015increasing} future development should expand their applied filter to include structured EHR data, such as vital signs and lab results, and better extract information from narrative criteria. \\

The future research of study \cite{beck2020artificial} should try to work on making the system applicable for a wider variety of cancers, integrating clinical data so the model can be better interpretable, and on whether such AI-driven systems might affect trial recruitment or the outcome of patients across a diversity of settings. For future work of paper \cite{zhang2017automated} some other studies could explore the use of these classifiers in real time, such as in search engines or patient trial matching systems, and observe how it impacts clinical decisions. Testing the approach on diseases such as diabetes or hypertension can also help actually see how it applies to a broader level. The research \cite{kang2017eliie}  could improve the ability of the system to identify clinical relationships and process complex queries by adding relational logic. Further key steps toward increasing applicability and scalability in different clinical settings would involve expanding the system to other diseases and testing its portability across different text sources.\\This study \cite{b9} has few limitations that point to future opportunities for research. Future work will refine the EHR-based "target patient" representation using these advanced techniques, try more complex statistical models, and test the methodology on trials involving other diseases to enhance generalizability and effectiveness. Future work of paper \cite{b15} will involve a comprehensive evaluation of the LCT corpus, possibly by using accuracy of query generation and semantic representation as metrics, which can use ROUGE scoring or UMLS Concept identifiers. Another good step toward that would be testing generated query methods on real clinical trial data from the University of Washington's EHR system for actual trial participants and thus for comparison.

\end{document}